\documentclass[letterpaper, 10 pt, conference]{ieeeconf}

\IEEEoverridecommandlockouts  

\let\labelindent\relax
\usepackage[utf8]{inputenc}
\usepackage[nolist, nohyperlinks, printonlyused, withpage]{acronym}
\usepackage{algorithm}
\usepackage{algpseudocode}
\usepackage{amsmath}
\usepackage{amssymb}
\usepackage{booktabs}
\usepackage{color}
\usepackage{csquotes}
\usepackage{enumitem}
\usepackage[T1]{fontenc}
\usepackage{hyperref}
\usepackage{listings}  
\usepackage{mathtools}
\usepackage{regexpatch}
\usepackage[caption=false,lofdepth,lotdepth]{subfig}
\usepackage{stfloats}
\usepackage{textcomp}
\usepackage{times}
\usepackage{tabularx}
\usepackage{wrapfig}

\usepackage{color}
\definecolor{black}{gray}{0} 

\usepackage{tikz}
\usepackage{pgfplots}
\DeclareUnicodeCharacter{2212}{−}
\usepgfplotslibrary{groupplots,dateplot}
\usetikzlibrary{patterns,shapes.arrows}
\pgfplotsset{compat=newest}

\makeatletter
\newcommand\fs@betterruled{%
  \def\@fs@cfont{\bfseries}\let\@fs@capt\floatc@ruled
  \def\@fs@pre{\vspace*{7pt}\hrule height.8pt depth0pt \kern2pt}%
  \def\@fs@post{\kern2pt\hrule\relax}%
  \def\@fs@mid{\kern2pt\hrule\kern2pt}%
  \let\@fs@iftopcapt\iftrue}
\floatstyle{betterruled}
\restylefloat{algorithm}
\makeatother

\newcommand*\Let[2]{\State #1 $\gets$ #2}  

\definecolor{maroon}{rgb}{0.5,0,0}
\definecolor{deepgreen}{rgb}{0,0.5,0}
\definecolor{deepblue}{rgb}{0,0,0.5}
\definecolor{deepred}{rgb}{0.6,0,0}
\definecolor{deepgreen}{rgb}{0,0.5,0}
\DeclareFixedFont{\ttb}{T1}{txtt}{bx}{n}{7} 
\DeclareFixedFont{\ttm}{T1}{txtt}{m}{n}{7}  
\lstdefinelanguage{XML}
{
  basicstyle=\ttfamily\footnotesize,
  morestring=[s]{"}{"},
  morecomment=[s]{?}{?},
  morecomment=[s]{!--}{--},
  commentstyle=\color{deepgreen},
  moredelim=[s][\color{black}]{>}{<},
  moredelim=[s][\color{red}]{\ }{=},
  identifierstyle=\color{maroon}
}

\lstdefinelanguage{python}
{
  basicstyle=\ttfamily\footnotesize,
  morekeywords={self, from, import},              
  morestring=[s]{'}{'},
  keywordstyle=\ttb\color{deepblue},
  emph={ModulesDB, ModuleAssembly},          
  emphstyle=\ttb,    
  stringstyle=\color{deepgreen},
  frame=tb,                         
  showstringspaces=false,
}
\lstdefinelanguage{bash}
{
  basicstyle=\ttfamily\footnotesize,
  morestring=[s]{'}{'},
  stringstyle=\color{deepgreen},
  showstringspaces=false,
}

\date{}

\makeatletter
\def\bstctlcite#1{\@bsphack
    \@for\@citeb:=#1\do{%
    \edef\@citeb{\expandafter\@firstofone\@citeb}%
    \if@filesw\immediate\write\@auxout{\string\citat
    ion{\@citeb}}\fi}%
\@esphack}
\makeatother

\title{\LARGE \bf
Timor Python: A Toolbox for Industrial Modular Robotics
}

\author{%
Jonathan K\"ulz,
Matthias Mayer,
and Matthias Althoff%
\thanks{All authors are with the Cyber-Physical Systems Group, Department of Computer Science, Technical University of Munich, 85748 Garching, Germany {\tt\small [jonathan.kuelz, matthias.mayer, althoff]@tum.de}.
Matthias Althoff and Jonathan K\"ulz are also with the Munich Center for Machine Learning (MCML).%
}}

\begin{document}
\bstctlcite{IEEEcustom:BSTcontrol}

\makeatletter

\begin{acronym}
    \acro{api}[API]{application programming interface}
    \acro{modrob}[MRR]{Modular Reconfigurable Robot}
    \acro{ga}[GA]{genetic algorithm}
    \acro{tool}[Timor]{Toolbox for Industrial Modular Robotics}
    \acro{URDF}{Unified Robot Description Format}
\end{acronym}
\maketitle

\thispagestyle{empty}
\pagestyle{empty}

\begin{abstract}
\noindent
\acp{modrob} represent an exciting path forward for industrial robotics, opening up new possibilities for robot design.
Compared to monolithic manipulators, they promise greater flexibility, improved maintainability, and cost-efficiency.
However, there is no tool or standardized way to model and simulate assemblies of modules in the same way it has been done for robotic manipulators for decades.
We introduce the \acf{tool}, a Python toolbox to bridge this gap and integrate modular robotics into existing simulation and optimization pipelines.
Our open-source library offers model generation and task-based configuration optimization for \acp{modrob}.
It can easily be integrated with existing simulation tools -- not least by offering URDF export of arbitrary modular robot assemblies.
Moreover, our experimental study demonstrates the effectiveness of Timor as a tool for designing modular robots optimized for specific use cases.

\end{abstract}


\section{Introduction}

\acfp{modrob} offer a promising extension to conventional robotic manipulators.
Assembled from a combination of individual hardware parts referred to as modules, they pose unprecedented flexibility, easy maintainability, and transportability by design \cite{Yim2007, Liu2016}.
However, the vast number of possible robot morphologies resulting from a given set of modules comes with significant challenges when designing \acp{modrob}~\cite{Paredis1993}.
Every configuration of modules leads to a new dynamic and kinematic model, which must be generated to apply control and planning procedures designed for traditional manipulators.
While there are multiple well-established frameworks for conventional robots enabling physical simulation, none of them can be used in a scenario where the robot model is subject to constant change.

\subsection{Contributions}
The \textbf{T}oolbox for \textbf{I}ndustrial \textbf{Mo}dular \textbf{R}obotics (Timor) provides modeling and simulation capabilities for \acp{modrob}, starting from a standardized module description.
Assembled configurations of modules can easily be exported to the \acf{URDF}\footnote{\url{https://wiki.ros.org/urdf}} for easy integration into existing pipelines.
Furthermore, it facilitates the design of task-tailored \acp{modrob} by incorporating search heuristics and optimization algorithms.
User-defined as well as provided module libraries can easily be assembled to robots with the possibility to automatically generate dynamic, kinematic, and collision models, all within the same framework.
Furthermore, any module combination -- also referred to as \textit{assembly} -- comes with visualization capabilities.
Timor is fully implemented in Python and hosted on \mbox{{\small \url{https://gitlab.lrz.de/tum-cps/timor-python}}}.
Code coverage, unit tests, documentation, and a moderated issue board are publicly available.
The toolbox and all dependencies can be installed with
\begin{lstlisting}[
    language=bash,
    tabsize=2,
    showstringspaces=false
]
$ pip install timor-python
\end{lstlisting}

\subsection{Related Work}

\acs{tool} enables modeling, simulating, and designing arbitrary configurations of robot modules, effectively bridging the divide between modular robots and traditional manipulators.
As a result, recent advancements in robotic simulation suites are now within reach for \acp{modrob}, taking an essential step toward their implementation in industrial settings.

\paragraph{\acfp{modrob}}
An \acs{modrob} consists of multiple independent modules that can be reconfigured externally or by the robot (self-reconfiguring modular robots).
The separation of module-level and system-level design enables rapid change-over, expansion, and robustness \cite{Yim2007, GuilinYang2022Intro}.
Early publications on the concept of \acp{modrob}, such as the CEBOT \cite{Fukuda1988} and the RMMS \cite{Schmitz1988} in 1988, already established a distinction between joint and link modules with the possible extension of functional (i.e., end effector) modules, a classification that is frequently used until today.
Over the last decades, various modular robot systems have been proposed \cite{Liu2016, Yun2020, Romiti2022}.
However, it is only recently that \acp{modrob} have finally left research laboratories.

The benefits of \acp{modrob} become particularly noteworthy when multiple tasks, which may not be known during the hardware design phase, need to be accomplished using a single robot or set of modules only \cite{Yim2007}.
However, utilizing modular robots in varying configurations often requires the expertise of professional and experienced programmers, rendering industrial implementation impractical.
This limitation has been overcome with the introduction of self-programming capabilities for modular robots \cite{Althoff2019}.
In recent years, multiple industrial solutions for modular robots entered the market\footnote{\url{https://www.hebirobotics.com/}}\textsuperscript{,}\footnote{\url{https://www.robco.de/en}} or have been announced for a future release\footnote{\url{https://www.beckhoff.com/en-en/products/motion/atro-automation-technology-for-robotics/}}.
Despite the increased supply in \acs{modrob} hardware and the acknowledgment of the necessity of fast model generation \cite{IMingChen1999}, distinct modeling libraries for industrial \acp{modrob} are not yet available and development and research departments often resort to tools designed for traditional robotics.

Theoretical groundwork for \acs{modrob} model generation have been published in recent years:
The CoBRA benchmark \cite{Mayer2022} provides a standardized module description format and benchmarks to compare \acp{modrob} and the authors of \cite{Bordignon2010, Bordignon2011} propose a language to describe \acp{modrob} through constraints implicitly.
Despite introducing a framework for an \acs{modrob} modeling language, these works lack the possibility of kinematic and dynamic model generation for explicitly defined configurations of robot modules.
Modular robot URDF files can be generated using a SolidWorks plugin as described in the work in \cite{Feder2022}.
However, its limitations include dependence on third-party software and a need for pre-defined CAD models.
The work in \cite{Nainer2021} proposes a procedure for automatic robot model description generation but only supports serially connected and asymmetric modules.

\acs{tool} extends existing theoretical works by offering support of kinematic trees and a unified module description format to model modules with an arbitrary number of connection interfaces, bodies, and joints.
It further implements this approach and leverages it for \acs{modrob} design, making it accessible for future research.

\paragraph{Search for \acs{modrob} configurations}
Identifying an optimized configuration of \acs{modrob} modules to meet desired robot performance characteristics poses a considerable challenge.
An exhaustive algorithm that uses enumeration of kinematically distinct configurations is theoretically applicable \cite{Chen1998}; however, the large search space renders this method impractical.
Various attempts have been made to simplify this problem:
In \cite{Althoff2019, Romiti2021}, the search is limited to configurations with constraints on the alternation between static links and joints.
The work in \cite{Liu2020} restricts the search even further to kinematics known from industrial robots.
In addition to a search space reduction, heuristic algorithms are commonly applied.
The early work in \cite{Sims1994} co-adapted morphology and control of virtual creatures using \acfp{ga}.
In \acp{ga}, solution candidates (\textit{chromosomes}) composed of a fixed number of variables (\textit{genes}) are evolved for a fixed number of generations by selection, reproduction, and mutation operations \cite{Goldberg1988}.
In applications of \acs{modrob} design, a representation of modules as genes and assemblies as chromosomes is usually utilized.
This approach has ever since been explored in various works \cite{Lipson2000, Icer2017, Wang2019}.
Furthermore, learning-based methods have recently been proposed to optimize modular agents both in industrial and non-industrial contexts \cite{Whitman2020, Zhao2020, Hu2022}.
While these approaches share a common robot design principle and are evaluated in simulation, they are based on individual software interfaces.
There is no possibility to evaluate them on a common set of robot modules, nor are they directly applicable to custom industrial tasks.
Timor provides a unified interface for constructing, evaluating, and optimizing \acs{modrob} models, allowing one to utilize arbitrary optimization algorithms in a plug-and-play manner.

\paragraph{Robotics Simulation}
Simulation tools offer great potential for developing control, morphology, and motion planning in robotics \cite{Choi2021}.
There is a wide range of sophisticated physics engines like \mbox{Bullet \cite{coumans2021}}, \mbox{MuJoCo \cite{Todorov2012}}, \mbox{DART \cite{Lee2018}}, or the Gazebo \mbox{Simulator \cite{Koenig2004}} that can be used to develop and fine-tune model-based controllers and precisely simulate robot and contact dynamics.
On the other side of the spectrum are toolboxes with a more narrow focus on kinematic and dynamic modeling in robotics, such as the Robotics Toolbox \cite{Corke2021} or Pinocchio \cite{JustinCarpentier2019}.
These tools, frequently used for tasks such as path planning and controller synthesis, commonly use a higher level of abstraction that realizes fast computation and provide a high-level \acf{api}.
While there is a great choice of tools for monolithic robots, all assume a known robot model.
A lack of support for \acp{modrob}, whose robot model is subject to change, requires users to craft workarounds for established simulation tools, mostly tailored to specific hardware.
Simulators like the USSR \cite{Christensen2008}, ReBots \cite{Collins2016}, or VisibleSim \cite{Thalamy2022} enable the simulation of self-reconfigurable robots that consist of up to millions of modules.
However, these simulators do not provide a high-fidelity computation of robot kinematics and dynamics, which is essential for achieving realistic simulations of industrial tasks.
There have been efforts to connect disassembled robot modules with traditional robotics libraries, but they are only applicable to specific sets of modules, such as the HEBI HRDF format\footnote{\url{https://github.com/HebiRobotics/hebi-hrdf}}.
\acs{tool} aims to provide a standardized interface to enable the modeling of arbitrary module libraries, making it possible to use established tools for \acp{modrob}.

\section{Model Generation}

Timor provides automatic model generation for module configurations, including a URDF export interface.
In addition, the models are used by the built-in simulation capabilities of Timor to support the design process of \acp{modrob}.

\subsection{Graph Representations}
Timor can compute two graph representations for robot modules and the relations between them.
They are generated automatically without additional user input and serve as an abstract and domain-independent view of \acp{modrob}.
A short overview of the terminology is given in the following -- details on the definition and generation of the graphs are provided in the referenced sections.
\begin{itemize}
    \item The directed \textit{module graph} $G_m$ (Sec. \ref{sec:Module}) represents a single module and, on its edges, contains information about the relative transformations between the reference frames of its constituent joints, bodies, and connectors.
    \item The directed \textit{assembly graph} $G_a$ (Sec. \ref{sec:kinematic_model}) is a combination of multiple module graphs. It represents the relative placement of every joint, body, and connector reference frame within an assembly of modules.
\end{itemize}
As a consequence of declaring the graphs directed, homogeneous transformations between their constituent elements can directly be assigned as edge features.

\subsection{Module Definition}\label{sec:Module}

\begin{figure*}
    \centering
    \begin{minipage}{0.49\linewidth}
        \centering
        \subfloat[Powerball module with reference frames $F$ for connectors ($F_{C1}$ and $F_{C2}$), bodies ($F_{B1}$, $F_{B2}$, and $F_{B3}$), and revolute joints ($F_{J1}$ and $F_{J2}$).\label{fig:Powerball_frames}]{
            \includegraphics[width=\linewidth]{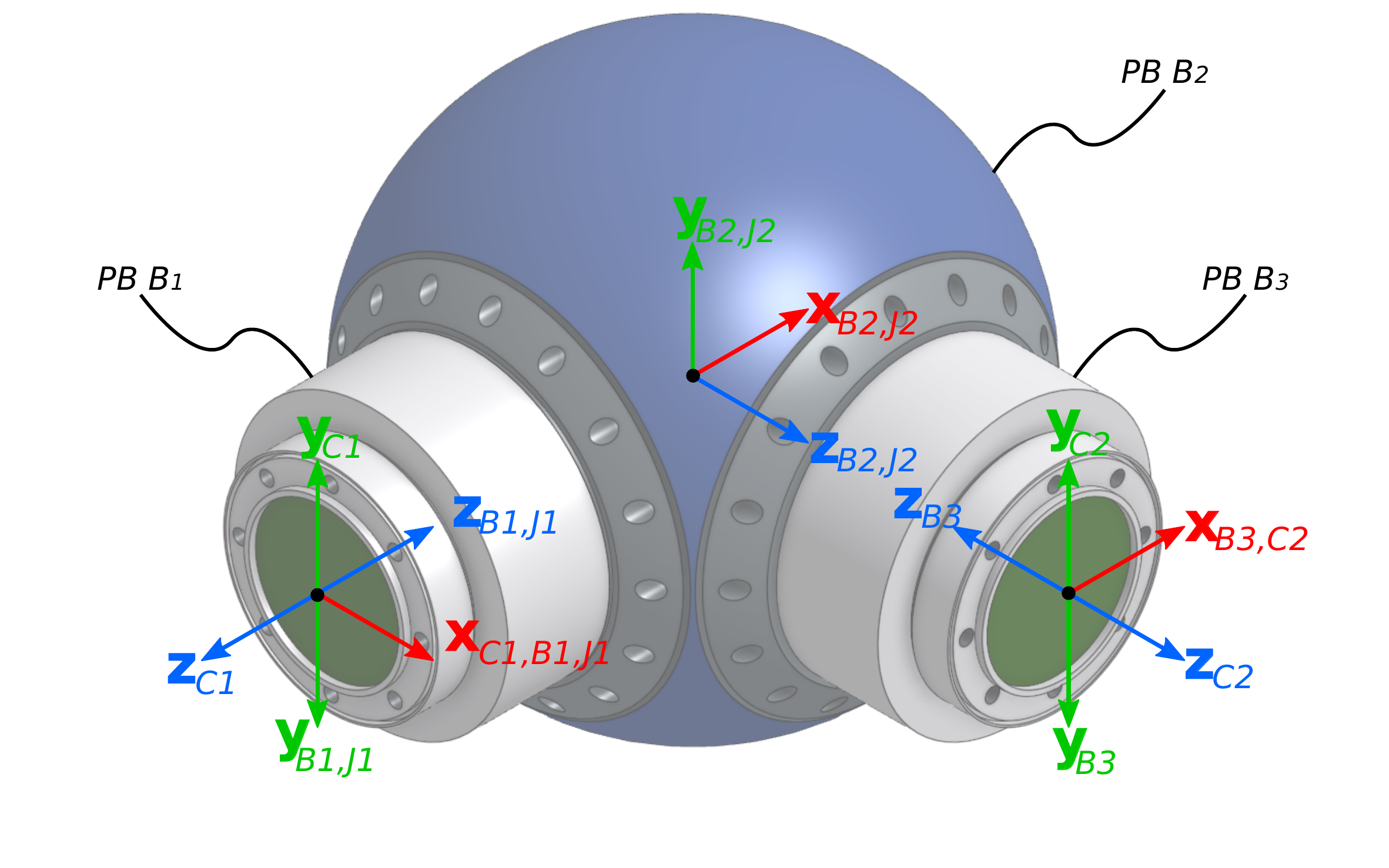}
            \hfill
        }
        \\
        \vspace{.66cm}
        \subfloat[Module graph $G_m$ for the Powerball module. The nodes of the graph represent bodies ($B_i$), joints ($J_i$), or connectors ($C_i$). The homogeneous transformations between their reference frames are stored on the edges of the graph.\label{fig:Powerball_graph}]{
            \includegraphics[width=\linewidth]{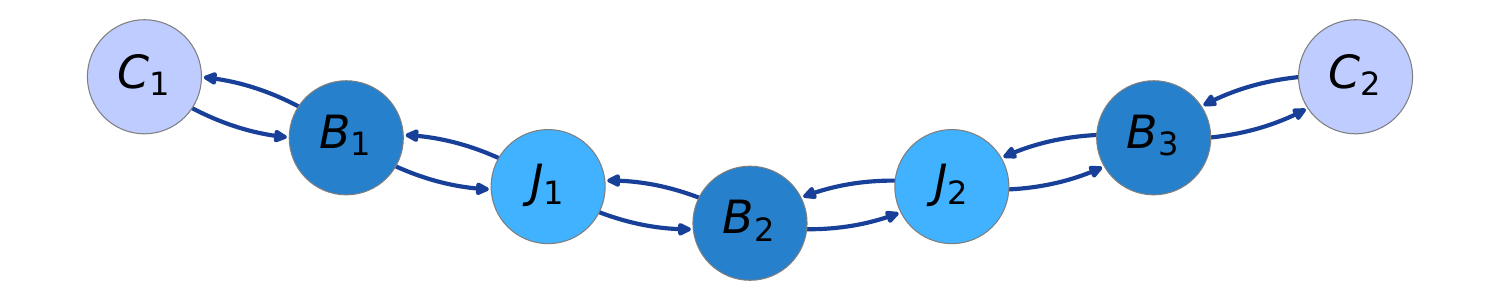}
            \hfill
        }
        \hfill
    \end{minipage}
    \hfill
    \begin{minipage}{0.49\linewidth}
        \centering
        \subfloat[The URDF excerpt shows the structure for links and joints modeling the Powerball module in an assembly, assuming it is attached to a base body.\label{fig:Powerball_urdf}]{
            \input{Powerball.tex}
            \hfill
        }
    \end{minipage}
    \caption[Module]{Relative placements of bodies, connectors, and joints in a module are defined by the positioning of their reference frames. The homogeneous transformations between reference frames are stored in the module graph $G_m$ that also captures neighbor relations. This information can be used to auto-generate URDF models for assemblies.}
    \label{fig:module}
\end{figure*}%

The implementation of modules in \acs{tool} is inspired by the framework introduced for the CoBRA benchmark \cite{Mayer2022}.
Modules are composed of rigid bodies and joints.
Bodies can have an arbitrary number of connectors, defining interfaces to attach them to other modules.
Every body, joint, and connector is assigned a reference frame $F$.
The placement and orientation of any joint or connector reference frame $F_i$ is defined relative to a body reference frame $F_j$ by a homogeneous transformation $T(F_i, F_j)$ -- in this case, we say there exists a neighbor relationship between the corresponding elements.
Fig.~\ref{fig:Powerball_frames} shows a module designed for the Schunk LWA 4P robot, consisting of three bodies, two connectors, two revolute joints, and the corresponding reference frames.

For any module $m_i$ with connectors $\mathcal{C}_{m_i}$, bodies $\mathcal{B}_{m_i}$, and joints $\mathcal{J}_{m_i}$, we define the directed \textit{module graph} as a tuple \mbox{$G_{m_i} = (\mathcal{V}_{m_i}, \mathcal{E}_{m_i})$} with
\begin{align*}
    \mathcal{V}_{m_i} & = \mathcal{C}_{m_i} \cup \mathcal{B}_{m_i} \cup \mathcal{J}_{m_i} \\
    \mathcal{E}_{m_i} & = \{(u, v) \in \mathcal{V}_{m_i} \times \mathcal{V}_{m_i} \mid \mathtt{neighbor}(u, v)\},
\end{align*}
where the $\mathtt{neighbor}(u, v)$ predicate evaluates to true if a neighbor relationship between $u$ and $v$ exists and false, otherwise.
To any edge $e \in \mathcal{E}_{m_i}$ between vertices $(u, v)$ with reference frames $(F_u, F_v)$, we assign the feature \mbox{$T_e \vcentcolon= T(F_u, F_v)$}.
Fig.~\ref{fig:Powerball_graph} shows the module graph for the Powerball module displayed in Fig.~\ref{fig:Powerball_frames}.%

While a module can be composed of an arbitrary number of these base elements, it usually represents one piece of hardware as produced by a manufacturer.
Due to the possible multiplicity of bodies and joints, even complex hardware with arbitrary geometries, such as fully integrated multiple degrees-of-freedom joints or mobile bases can easily be described as a single module.

\subsection{Connectors}
Connectors are interfaces placed on bodies, used to define a connection between two modules.
Any connector $C_i$ has a \mbox{\textit{gender} $g_i \in$ \{\textbf{m}ale, \textbf{f}emale, \textbf{h}ermaphroditic\}}, a \textit{type} $t_i$, and a \textit{size} $s_i$.
There are two special connector types, \textit{eef} and \textit{base}, that define end effector frame(s) and the base reference frame(s).
Apart from these reserved keywords, other types can be chosen freely by users to reflect the connector hardware (e.g., \textit{flange}, \textit{clamp}).
We define the compatibility function between two connectors $(C_i, C_j)$ as
\begin{equation*}
\sigma_c(C_i, C_j) \vcentcolon=
\begin{cases}
\text{true,} &
    \begin{aligned}
        &\hspace{-0.7em}\text{if } t_i = t_j, \\
        &\hspace{-0.7em}\text{and }s_i = s_j, \\
        &\hspace{-0.7em}\text{and}\begin{cases}
            g_i \neq g_j \text{, if }g_i, g_j \in \{m, f\}, \\
            g_i = g_j = h \text{, otherwise}.
        \end{cases}
    \end{aligned} \\
\text{false,} &\hspace{-0.7em}\text{otherwise}.
\end{cases}
\end{equation*}
Similarly, we define the compatibility function between two modules $(m_1, m_2)$ with connectors $(\mathcal{C}_{m_1}, \mathcal{C}_{m_2})$ as
\begin{equation*}
    \sigma_m(m_1, m_2) \vcentcolon=
    \begin{cases}
        \text{true,} & 
        \hspace{-1em}\text{if } \exists (C_i, C_j)\,{\in}\,\mathcal{C}_{m_1}\!{\times}\,\mathcal{C}_{m_2}{\vcentcolon}\,\sigma_c(C_i, C_j) \\
        \text{false,} & \hspace{-0.7em}\text{otherwise.}
    \end{cases}
\end{equation*}
The possibility to assign an arbitrary number of connectors to a body and the introduction of hermaphroditic connectors allows one to model non-serial kinematics and (multidirectional) modules without a unique mounting orientation.
Without loss of generality, we assume that the reference frames $(F_{i}, F_{j})$ of the connectors $(C_i, C_j)$ when connected are arranged such that the x-axes are aligned, and the z-axes are pointing away from the module (Fig. \ref{fig:Powerball_frames}).
Therefore, \mbox{$T(F_{i}, F_{j}) = R_x(\pi)$} where $R_x(\pi)$ is a homogeneous transformation performing a rotation of 180° around the x-axis.

\subsection{Assembling Modules}\label{sec:module_sets}
We denote a set of robot modules as $\mathcal{R}$.
A configuration of modules \mbox{$M = \{m_1, \dots, m_n \}$} with connections \mbox{$\Sigma = \{\{C_a, C_b\}, \{C_c, C_d\}, \dots \}$} is referred to as an assembly $\mathcal{A}$.
Timor efficiently depicts both serial and branched kinematics.
For representing closed-chain kinematics, Timor provides a robust framework, although simulation and validity checks require utilization of external libraries.
Recognizing that many \acs{modrob} configurations are chains, Timor also allows specifying an assembly implicitly as a tuple of modules $(m_1, m_2, \dots)$ as long as there is exactly one possible connection between any two neighboring modules $(m_i, m_{i+1})$.

\begin{figure}
    \centering
    \includegraphics[width=\linewidth]{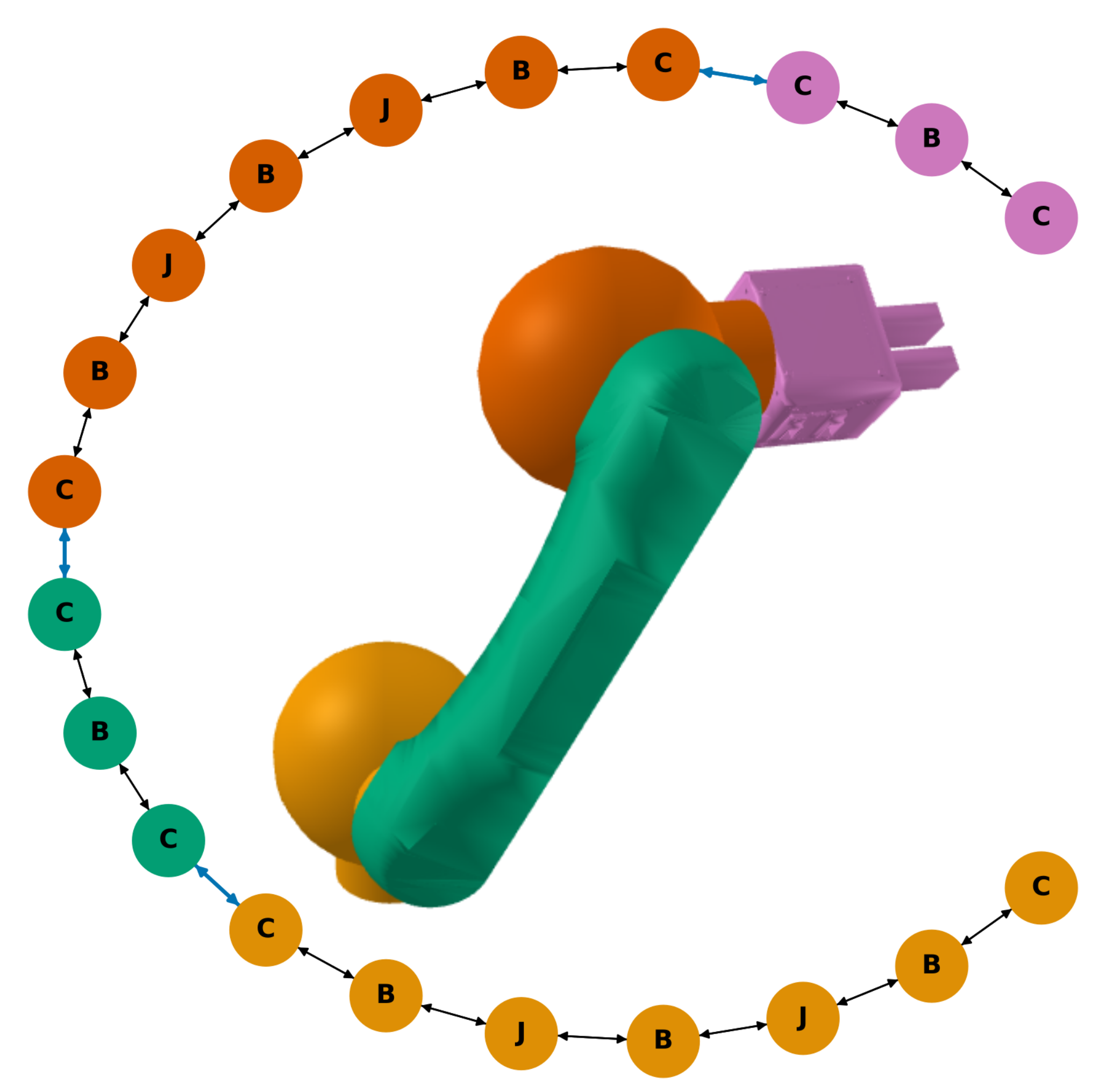}
    \caption[Assembly Graph]{A Schunk LWA 4p robot built from four modules and the corresponding assembly graph $G_a$. The nodes of the composing module graphs $G_m$ are drawn in the same color as the corresponding modules. Edges connecting modules ($\Sigma_d$) are highlighted in blue.}
    \label{fig:schunk_ga}
\end{figure}
\subsection{Building Kinematic and Dynamic Models}\label{sec:kinematic_model}
For deriving the kinematic and dynamic model of an assembly, we introduce the directed \textit{assembly graph} \mbox{(Fig. \ref{fig:schunk_ga})} as the tuple $G_a = (\mathcal{V}_a, \mathcal{E}_a)$ with
\begin{align*}
    \mathcal{V}_a & = \mathcal{V}_{m_1} \cup \dots \cup \mathcal{V}_{m_n}, \\
    \mathcal{E}_a & = \Sigma_d \cup \mathcal{E}_{m_1} \cup \dots \cup \mathcal{E}_{m_n},
\end{align*}
where $\Sigma_d$ represents the connections between different modules and contains, for every connection in $\Sigma$, two directed, antiparallel edges \mbox{(Fig. \ref{fig:schunk_ga})} to which we assign the feature \mbox{$T_e = T_e^{-1} \vcentcolon= R_x(\pi)$}.

For any given sequence \mbox{$E(u, v) = (e_1, \ldots, e_n)$} of edges on a path between two nodes $(u, v)$ in $G_a$, the homogeneous transformation between any two reference frames $T(F_u, F_v)$ is
\begin{equation*}
    T(F_u, F_v) = T_{e_1} T_{e_2} \ldots T_{e_n}, \qquad u \neq v
\end{equation*}
Obviously, any sequence $E$ works; we usually determine the shortest one using breadth-first search.

\begin{algorithm}[t!]
\caption{Generate URDF}
\label{alg:parse_urdf}
\begin{algorithmic}[1]
    \Function{generate\_urdf}{$Assembly, name$}
    \Let{$G$}{$Assembly$ graph $G_a$}
    \Let{$C_b$}{$Assembly$ base connector}
    \Let{$urdf$}{initialize URDF($name$)}
    \Let{$T$}{$\{C_b.frame: \mathbb{I}_4\}$} \Comment{$\mathbb{I}_4 \gets$identity matrix}
    \For{(node $n$, edge $e$, successor $s$) $\gets$ bfs($G, C_b$)}
        \Let{$F_n, F_s$}{$n.frame, s.frame$}
        \Let{$T_s$}{$T[F_n] \cdot e.T$}
        \Let{$T[F_s]$}{$T_s$}
        \If{is body($s$)}
            \Let{$link$}{make URDF link ($s, T_s$)}
            \Let{$urdf$}{append($urdf, link$)}
        \ElsIf{is joint($s$)}
            \Let{$joint$}{make URDF joint ($s, T_s$)}
            \Let{$urdf$}{append($urdf, joint$)}
            \Let{$T[F_s]$}{$\{F_b: \mathbb{I}_4\}$}
        \ElsIf{is connector($s$)}
            \Let{$joint$}{make URDF fixed joint ($s, T_s$)}
            \Let{$urdf$}{append($urdf, joint$)}
            \Let{$T[F_s]$}{$\{F_b: \mathbb{I}_4\}$}
        \EndIf
    \EndFor
    \EndFunction
\end{algorithmic}
\end{algorithm}
By performing a breadth-first iteration over $G_a$, starting at the base reference frame, \acs{tool} generates URDF descriptions for arbitrary module arrangements, as shown in Alg. \ref{alg:parse_urdf}, as long as $G_a$ composes a tree (URDF does not natively support closed-chain kinematics).
Homogeneous transformations from parent joints to URDF elements are stored in the map $T$, so $T[F]$ is the relative transformation between a frame and its parent joint.
All further required information to write a corresponding URDF element\footnote{URDF links: \url{https://wiki.ros.org/urdf/XML/link}}\textsuperscript{,}\footnote{URDF joints: \url{https://wiki.ros.org/urdf/XML/joint}} such as inertia or joint limits are stored within the modules.
Fig.~\ref{fig:Powerball_urdf} shows the resulting structure for a partial URDF generated for the module shown in Fig.~\ref{fig:Powerball_frames}.

The generation of a dynamic and kinematic model from URDF that can be used for simulation can be achieved by various software packages, such as the open-source tool Pinocchio \cite{JustinCarpentier2019, pinocchioweb}.
For the sake of computational efficiency, we also provide a direct conversion from any assembly to Pinocchio, omitting an intermediate URDF generation.
Furthermore, \acs{tool} offers an interface to the FCL library \cite{Pan2012} that we use for efficient collision checking.

\section{MRR configuration search}
As the number of possible \acs{modrob} configurations grows exponentially with the number of assembled modules, the search for an optimal configuration for an application cannot be performed exhaustively in general.
Timor offers multiple tools to aid human and algorithmic optimization of assemblies for robot tasks.

\subsection{Task Definition}
Tasks can be formally described in the format introduced in \cite{Mayer2022}.
They are composed of goals that can be end-effector poses to reach or trajectories to follow.
All tolerances, obstacles, and constraints supported by the CoBRA benchmark can be specified and visualized with Timor, thus supporting a wide range of applications.

\subsection{Visualization}
\begin{figure}
    \vspace{7pt}  
    \begin{lstlisting}[
    language=python,
    ]
import numpy as np
from timor.Module import *
from timor.utilities.visualization import animation

db = ModulesDB.from_json_file(db_file)
modules = ('base', 'J2', 'i_45', 'J2', 'J2', 'eef')
A = ModuleAssembly.from_serial_modules(db, modules)
q0 = A.robot.random_configuration()
q1 = A.robot.random_configuration()
trajectory = np.linspace(q0, q1)
animation(A.robot, trajectory, dt=.1)
    \end{lstlisting}
\caption{Code sample to generate an animation.}
\label{fig:movie_sample}
\end{figure}
Timor extends the meshcat\footnote{\url{https://github.com/rdeits/meshcat-python}} visualizer capabilities that are integrated in Pinocchio to enable manual inspection in an interactive browser-based visualization for robots and tasks.
Furthermore, \acs{tool} allows generating videos for robot trajectories and exporting them as a file.
Fig.~\ref{fig:movie_sample} shows a code sample for loading a set of modules, defining an assembly, generating the corresponding robot model, and visualizing a random trajectory.

\subsection{Pruning Iterators}
Timor can enumerate valid combinations between modules and can produce configurations satisfying human-specified constraints without the need of pre-computing all valid configurations, therefore being memory-efficient.
In particular, reasonable alternations between static links and joints, as proposed in \cite{Althoff2019, Romiti2021}, can be enforced -- not at least to resemble kinematics from industrial robots as in \cite{Liu2020}.
Furthermore, a custom set of valid sequences of modules can be defined explicitly, and the range of desired degrees of freedoms for an \acs{modrob} can be set.

\subsection{Assembly Filters}
In the search for task-specific \acp{modrob}, the feasibility of sub-problems is often checked.
Simple metrics can be used to eliminate a majority of unfit \acs{modrob} morphologies before performing computationally expensive evaluations~\cite{Icer2017}.
Timor implements \textit{filters} on criteria that are first evaluated for the goal pose and later for an entire trajectory, such as
\begin{itemize}
    \item kinematic solutions exist,
    \item no self-collision, or
    \item no collision with the environment.
\end{itemize}
By evaluating metrics prior to model generation whenever feasible, the benefits of having a comprehensive and integrated library are demonstrated.

\subsection{Genetic Algorithms (GAs)}
The flexibility of \acp{modrob} can be leveraged to mimic various industrial robot kinematics.
However, overly strong constraints on the search space for \acs{modrob} configurations can limit the potential to discover specialized, previously unseen kinematics.
Metaheuristics, such as \acp{ga}, offer a potential solution by providing a method to explore a range of \acs{modrob} configurations while efficiently navigating the search space.
Timor provides a user-friendly interface to integrate \acp{ga} into the optimization process of \acs{modrob} configurations.
Users can define industrial tasks, load a set of available modules, and optimize their configuration using \acp{ga} within minutes.
As demonstrated in Sec. \ref{sec:user_study}, these capabilities can be utilized to uncover unique robot morphologies.

\section{Numerical Experiments}\label{sec:numerical_experiments}
To demonstrate the usefulness of Timor, we conducted two experiments.
In our first experiment, we use a module set that was published in \cite{Althoff2019} to generate URDF models for arbitrary combinations of modules, showing how Timor can be leveraged to extend and unify existing work on \acp{modrob}.
In our second experiment, we compare two optimization algorithms to human experts in designing a use-case-tailored \acs{modrob}.
The code to reproduce both experiments and more details on experimental results are provided together with a set of tutorials in the official repository\footnote{\url{https://gitlab.lrz.de/tum-cps/timor-python/-/tree/main/tutorials}}.
All experiments were conducted on a desktop PC with an Intel i7-11700KF processor.

\subsection{Model Generation for an Existing Module Set}
For this experiment, we use the IMPROV module set as introduced in \cite{Althoff2019}, which is composed of modules of the Schunk LWA 4P robot and additionally designed modules.
It contains two Powerballs (shown in Fig.~\ref{fig:Powerball_frames}), ten static links of varying sizes and shapes, and a base module.
By applying Timor's \textit{assembly iterator}, we obtain the 32,768 possible configurations of modules for a six degrees-of-freedom robot as reported in \cite{Althoff2019}.
Generating URDF files for all of them takes 127 seconds (3.9ms on average).
Directly transforming an assembly to a Pinocchio-based robot model is equally computationally expensive.
By providing tens of thousands of kinematic and dynamic models in a standardized format within a matter of minutes, Timor facilitates research on \acp{modrob}.

\subsection{User Study}\label{sec:user_study}
\begin{figure}
    \centering
    \includegraphics[width=\linewidth]{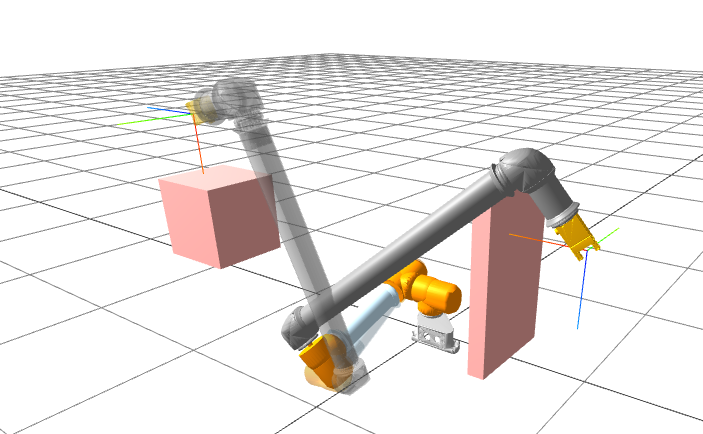}
    \caption[Best lightweight assembly]{Best lightweight assembly: The lightest robot reaching the two goals (marked as coordinate systems) while avoiding obstacles (red) was found by a \acs{ga}.}
    \label{fig:ga_optim}
\end{figure}

In a simulated environment, we challenged human experts to optimize the configuration of an \acs{modrob} given the task in Fig. \ref{fig:ga_optim} and compared their results to two algorithms in Timor.
A common use case for modular robots is machine tending, where the robot has to perform a pick-and-place operation in a potentially cluttered environment.
We abstract this task by fixing a robot base in an environment with two static obstacles and two goal poses that need to be reached.
An assembly poses a valid solution if an inverse kinematics solution can be found for both goals, given a position tolerance of 0.01mm and an orientation tolerance of 45° around an arbitrary axis relative to the desired positioning.
While path planning was not explicitly incorporated into the optimization process, we could identify valid trajectories for animating both the algorithm and human-generated results by adopting basic sampling-based planners.

\begin{figure}
    \centering
    \includegraphics[width=\linewidth]{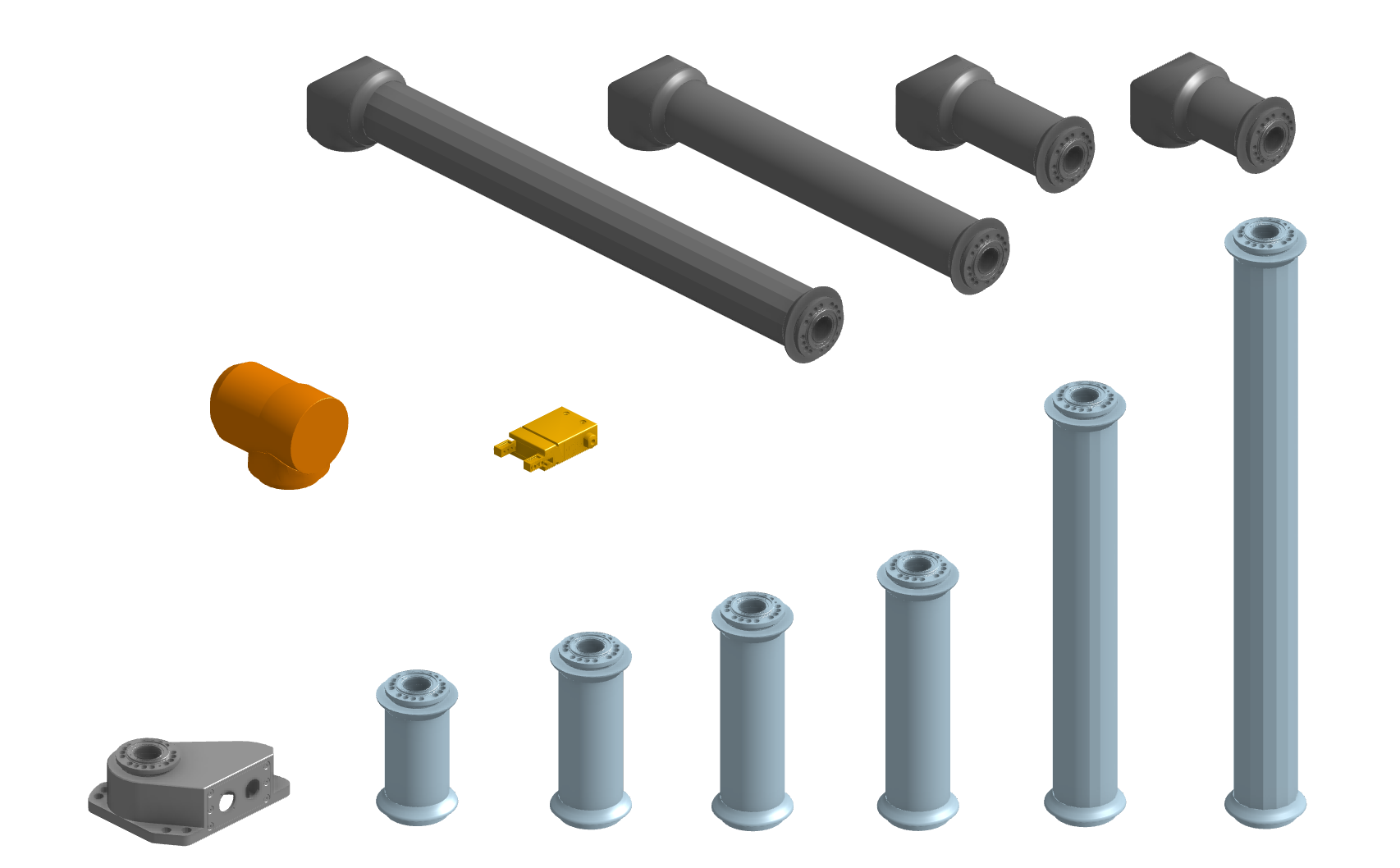}
    \caption[Modules of the user study]{Modules of the user study: For the experiment, we provide ten static link modules (grey) and one module containing a revolute joint (orange). The base and end effector are automatically added to every assembly.}
    \label{fig:robco_modules}
\end{figure}

For this experiment, we provide module data for robot modules manufactured by RobCo\footnote{\url{https://www.robco.de/en}} -- for the remainder of this section, we refer to modules from this set that contain a joint as \textit{joint modules} and static modules as \textit{base, end effector, and links}.
Fig.~\ref{fig:robco_modules} shows the module set used, consisting of four L-shaped links, six I-shaped links and one joint, base, and gripper module each.
The objective is to find an assembly that can reach both goals while minimizing the sum of individual module costs.
This can be a weighted sum of the acquisition cost, availability preferences, or module mass.
For this experiment, we assign costs directly proportional to the module mass.
We asked eight experts\footnote{Either people with an engineering degree focused on robotics or master students within the last year of a robotics-specific master program. All experts had prior experience working with \acp{modrob}.} to find an optimal solution to the task described above within 30 minutes (excluding briefing and debriefing).
After each guess, a user interface provides feedback on the cost and goal reachability of the chosen solution.
Furthermore, an animation is made available to the expert showing the assembly in the closest configuration to the goals found using a numerical inverse kinematics solver integrated in Timor.

We compare the human performance against a constrained search algorithm and a \acf{ga}.
The number of configurations achievable from a given set of modules can be limited through the utilization of Timor's pruning iterators;
for the constrained search, robots can have at most five degrees of freedom, at most one successive link between any two joint modules, the first module after the base must contain a joint, and there can be at most one link module between the last joint and the end effector.
Under these assumptions, there are
\begin{equation*}
    \sum_{n=1}^{dof} |J| \cdot ((|L| + 1) \cdot |J|)^n = \sum_{n=1}^{5} 11^n = 177,155
\end{equation*}
assemblies with exactly one base and end effector.
We integrate the built-in filters that prevent further assembly evaluation if a lighter solution has already been found or if, even without considering the obstacles, no inverse kinematics solution can be found that satisfies joint torque constraints.
By leveraging a comprehensive evaluation pipeline within a single library, we are able to pre-filter a large number of solutions prior to model generation.
This drastically increases efficiency, leading to a total runtime of 5.5 minutes.

Lastly, we utilize the Timor optimization interface and a genetic algorithm provided by the PyGAD library \cite{Gad2021}.
Each solution candidate is encoded as a chromosome consisting of thirteen genes that define the module configuration of the robot.
We limit the search space to alternating joint and link genes, where each gene can either represent a \{link, joint\} module (fig.~\ref{fig:robco_modules}) or an empty slot.
While restricting the search space to robots with at most six degrees of freedom, other than in the constrained search, an alternation of joints and links is not enforced due to the possibility of empty slots.
The solution space consists of
\begin{equation*}
    (|J|+1)^6 \cdot (|L|+1)^7 = 2^6 \cdot 11^7 = 1,247,178,944
\end{equation*}
different chromosomes -- some of which represent equivalent assemblies\footnote{The two chromosomes 'J-L1-0-L1-0-0-J1' and 'J-L1-0-0-0-L1-J' are different encodings for a chain of modules with IDs 'J-L1-L1-J'.}.
We perform 100 trials of 500 generations, each with an average runtime of 21.9 minutes per trial.
Fig. \ref{fig:ga_optim} shows the final solution found by the best trial.

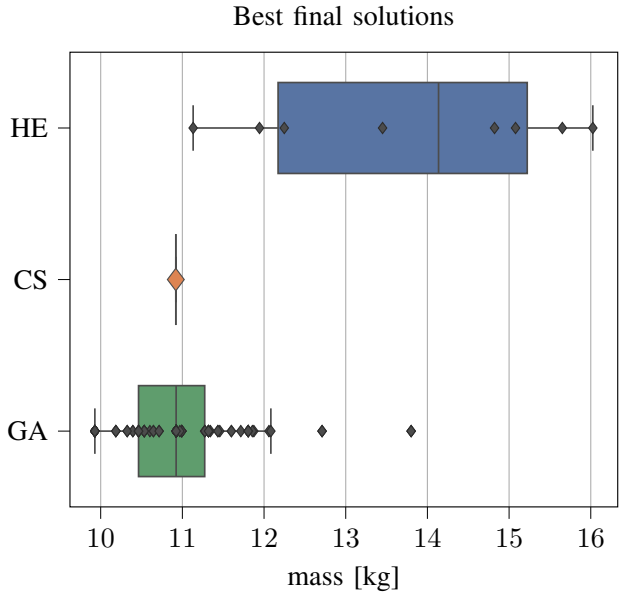
\begin{figure}
    \centering
    \resizebox{\columnwidth}{!}{
\begin{tikzpicture}

\definecolor{darkgray176}{RGB}{176,176,176}
\definecolor{darkslategray45}{RGB}{45,45,45}
\definecolor{darkslategray76}{RGB}{76,76,76}
\definecolor{dimgray90}{RGB}{90,90,90}
\definecolor{mediumseagreen95157109}{RGB}{95,157,109}
\definecolor{peru20313699}{RGB}{203,136,99}
\definecolor{peru22113282}{RGB}{221,132,82}
\definecolor{steelblue88116163}{RGB}{88,116,163}

\begin{axis}[
tick align=outside,
tick pos=left,
title={Best final solutions},
x grid style={darkgray176},
xlabel={mass [kg]},
xmajorgrids,
xmin=9.623907885, xmax=16.329998815,
xtick style={color=black},
y dir=reverse,
y grid style={darkgray176},
ymin=-0.5, ymax=2.5,
ytick style={color=black},
ytick={0,1,2},
yticklabels={HE,CS,GA}
]
\path [draw=darkslategray76, fill=steelblue88116163, semithick]
(axis cs:12.1719765,-0.3)
--(axis cs:12.1719765,0.3)
--(axis cs:15.222182825,0.3)
--(axis cs:15.222182825,-0.3)
--(axis cs:12.1719765,-0.3)
--cycle;
\path [draw=darkslategray76, fill=peru20313699, semithick]
(axis cs:10.92,0.7)
--(axis cs:10.92,1.3)
--(axis cs:10.92,1.3)
--(axis cs:10.92,0.7)
--(axis cs:10.92,0.7)
--cycle;
\path [draw=darkslategray76, fill=mediumseagreen95157109, semithick]
(axis cs:10.4635359,1.7)
--(axis cs:10.4635359,2.3)
--(axis cs:11.2735987,2.3)
--(axis cs:11.2735987,1.7)
--(axis cs:10.4635359,1.7)
--cycle;
\addplot [semithick, darkslategray76]
table {%
12.1719765 0
11.132083 0
};
\addplot [semithick, darkslategray76]
table {%
15.222182825 0
16.0251765 0
};
\addplot [semithick, darkslategray76]
table {%
11.132083 -0.15
11.132083 0.15
};
\addplot [semithick, darkslategray76]
table {%
16.0251765 -0.15
16.0251765 0.15
};
\addplot [semithick, darkslategray76]
table {%
10.92 1
10.92 1
};
\addplot [semithick, darkslategray76]
table {%
10.92 1
10.92 1
};
\addplot [semithick, darkslategray76]
table {%
10.92 0.85
10.92 1.15
};
\addplot [semithick, darkslategray76]
table {%
10.92 0.85
10.92 1.15
};
\addplot [semithick, darkslategray76]
table {%
10.4635359 2
9.9287302 2
};
\addplot [semithick, darkslategray76]
table {%
11.2735987 2
12.0835818 2
};
\addplot [semithick, darkslategray76]
table {%
9.9287302 1.85
9.9287302 2.15
};
\addplot [semithick, darkslategray76]
table {%
12.0835818 1.85
12.0835818 2.15
};
\addplot [black, mark=diamond*, mark options={solid,fill=darkslategray76}, only marks]
table {%
13.8008762 2
12.7104797 2
};
\addplot [semithick, darkslategray76]
table {%
14.1364983 -0.3
14.1364983 0.3
};
\addplot [semithick, darkslategray76]
table {%
10.92 0.7
10.92 1.3
};
\addplot [semithick, darkslategray76]
table {%
10.9237365 1.7
10.9237365 2.3
};
\addplot [draw=darkslategray45, fill=darkslategray76, mark=diamond*, only marks]
table{%
x  y
13.4510935 0
14.8219031 0
12.2477406 0
16.0251765 0
15.6519911 0
11.9446842 0
15.0789134 0
11.132083 0
};
\addplot [draw=darkslategray45, draw=none, fill=darkslategray76, mark=diamond*]
table{%
x  y
-3.33066907387547e-17 -0.707106781186547
0.424264068711928 0
2.22044604925031e-17 0.707106781186547
-0.424264068711929 1.11022302462516e-16
-3.33066907387547e-17 -0.707106781186547
};
\addplot [draw=darkslategray45, fill=darkslategray76, mark=diamond*, only marks]
table{%
x  y
11.4351395 2
9.9287302 2
11.3430475 2
10.5329847 2
10.3246383 2
10.3940872 2
9.9287302 2
11.8057865 2
11.2735987 2
11.3196446 2
10.9237365 2
9.9287302 2
10.9237365 2
10.9237365 2
11.8752354 2
9.9287302 2
10.9237365 2
10.9237365 2
11.7155527 2
10.9931854 2
10.5329848 2
13.8008762 2
9.9287302 2
10.9237365 2
10.9237365 2
10.9237365 2
11.3196446 2
10.9237365 2
9.9287302 2
11.8057865 2
11.2735987 2
11.8569889 2
10.9237365 2
10.9931854 2
11.3196447 2
10.9237365 2
10.4635359 2
10.6024336 2
10.6459412 2
10.9237365 2
10.9237365 2
11.4585423 2
11.600058 2
12.0601791 2
10.9237365 2
11.8057865 2
10.9237365 2
10.9237365 2
10.9931854 2
10.4635359 2
9.9287302 2
10.9931854 2
10.9237365 2
10.3940872 2
10.9237365 2
10.9237365 2
10.9237365 2
11.6000579 2
11.3196447 2
11.8057865 2
10.9931854 2
11.2761371 2
10.5329848 2
9.9287302 2
10.4635359 2
10.5329848 2
10.3246384 2
9.9287302 2
9.9287302 2
10.5329847 2
10.9237365 2
10.7153901 2
9.9287302 2
11.3196447 2
10.9237365 2
10.9237365 2
11.8057865 2
10.9237365 2
10.9237365 2
10.1857406 2
10.9237365 2
9.9287302 2
10.4635359 2
10.9237365 2
10.1857407 2
12.0835818 2
10.6459413 2
10.9237365 2
10.4635359 2
10.1857406 2
10.9237365 2
12.7104797 2
10.9237365 2
10.9237365 2
10.7153901 2
10.9697826 2
9.9287302 2
10.9237365 2
11.4351395 2
9.9287302 2
};
\addplot [draw=dimgray90, draw=none, fill=peru22113282, mark=diamond*]
table{%
x  y
-3.33066907387547e-17 -0.707106781186547
0.424264068711928 0
2.22044604925031e-17 0.707106781186547
-0.424264068711929 1.11022302462516e-16
-3.33066907387547e-17 -0.707106781186547
};

\addplot [draw=darkslategray76, fill=peru22113282, mark=diamond*, only marks, mark size=4pt]
table{%
x  y
10.92 1
};
\end{axis}

\end{tikzpicture}
    }
    \caption[User study results]{User study results: The best result (9.92kg) was found using a GA. While the median GA result (10.92kg) is equal to the constrained search (CS) result, none of the human experts (HE) was able to find a solution with a mass of less than 11kg (median: 14.1kg).}
    \label{fig:ga_history}
\end{figure}
One of the human experts found a robot with a total mass of 11.13kg, whereas the median expert designed a valid solution with a robot mass of 14.1kg (Fig. \ref{fig:ga_history}).
Within a fraction of the time, the constrained search yielded a robot with a mass of 10.92kg.
Leveraging the goal tolerances, the best trial of the \acs{ga} resulted in a three degrees-of-freedom robot with a total mass of 9.92kg, while even the worst trial ended with a solution with a total robot mass of 13.8kg (Fig. \ref{fig:ga_history}).
The best configuration identified by the \acs{ga} clearly is untypical (Fig. \ref{fig:ga_optim}): Three joints close to the base in combination with two static, L-shaped links provide enough flexibility for maneuvering between the goals while minimizing the total mass.

\section{Conclusion}
We introduced and showcased \acs{tool}, the first Python toolbox to model, simulate, and optimize modular reconfigurable robots.
Our open-source library is published under the MIT license and based on readily available components, offering easy integration in any pipeline without the need for commercial software.
The code examples and introductory tutorials are provided with the toolbox repository, along with the source code for all experiments conducted.
Bridging the simulation gap between conventional and modular robots, \acs{tool} offers features necessary for configuration optimization and model generation of \acp{modrob}.

\section*{Acknowledgment}
The authors gratefully acknowledge financial support by the
Horizon 2020 EU Framework Project CONCERT under grant 101016007 and by the
ZiM project on energy- and wear-efficient trajectory generation (ZF4086011PO8).

\addtolength{\textheight}{-12cm}   


\bibliographystyle{IEEEtranCustom}
\bibliography{bibstyle.bib, bibliography.bib}

\clearpage

\renewcommand\thefigure{A\arabic{figure}}
\setcounter{figure}{0}

\end{document}